\documentclass[letterpaper, 10 pt, journal, twoside]{IEEEtran}
\usepackage{cite}
\usepackage{amsmath,amssymb,amsfonts}
\usepackage{algorithmic}
\usepackage{graphicx}
\usepackage{textcomp}
\usepackage{bm}
\usepackage{multirow}
\usepackage[tight]{subfigure}
\usepackage{algorithm}
\usepackage{threeparttable}
\usepackage{hyperref}
\usepackage{xcolor}

\ifCLASSINFOpdf
\else
\fi

\begin{document}
%
\title{Transformer Driven Visual Servoing for Fabric Texture Matching Using Dual-Arm Manipulator}
%
%
%

\author{Fuyuki Tokuda$^{1}$, Akira Seino$^{1}$, Akinari Kobayashi$^{1}$, Kai Tang$^{1}$, and Kazuhiro Kosuge$^{1}$%
\thanks{Manuscript received: August, 1, 2025; Revised October, 22, 2025; Accepted November, 19, 2025.}
\thanks{This paper was recommended for publication by Editor Júlia Borràs Sol upon evaluation of the Associate Editor and Reviewers' comments.
This work was supported in part by the Innovation and Technology Commission of the HKSAR Government under the InnoHK initiative, in part by the JC STEM Lab of Robotics for Soft Materials funded by The Hong Kong Jockey Club Charities Trust, and in part by Tohoku University through the joint research program.} 
\thanks{$^{1}$Fuyuki Tokuda, Akira Seino, Akinari Kobayashi, Kai Tang, and Kazuhiro Kosuge are with the JC STEM Lab of Robotics for Soft Materials, Department of Electrical and Electronic Engineering, Faculty of Engineering, The University of Hong Kong, Hong Kong SAR, China, and also with the Centre for Transformative Garment Production, Hong Kong SAR, China
        {\tt\footnotesize fuyuki.tokuda@transgp.hk, akira.seino@transgp.hk; akinari.kobayashi@transgp.hk; tangkai@eee.hku.hk; kosuge@hku.hk}}%
\thanks{Digital Object Identifier (DOI): see top of this page.}
}
%
%

\markboth{IEEE Robotics and Automation Letters. Preprint Version. Accepted November, 2025}
{Tokuda \MakeLowercase{\textit{et al.}}: Transformer Driven Visual Servoing for Fabric Texture Matching Using Dual-Arm Manipulator}

%



\maketitle

\begin{abstract}
In this paper, we propose a method to align and place a fabric piece on top of another using a dual-arm manipulator and a grayscale camera, so that their surface textures are accurately matched. We propose a novel control scheme that combines Transformer-driven visual servoing with dual-arm impedance control. This approach enables the system to simultaneously control the pose of the fabric piece and place it onto the underlying one while applying tension to keep the fabric piece flat. Our transformer-based network incorporates pre-trained backbones and a newly introduced Difference Extraction Attention Module (DEAM), which significantly enhances pose difference prediction accuracy. Trained entirely on synthetic images generated using rendering software, the network enables zero-shot deployment in real-world scenarios without requiring prior training on specific fabric textures. Real-world experiments demonstrate that the proposed system accurately aligns fabric pieces with different textures.
\end{abstract}


%
\IEEEpeerreviewmaketitle

\section{Introduction}
\begin{figure}[t]
\centering
\includegraphics[width=68mm]{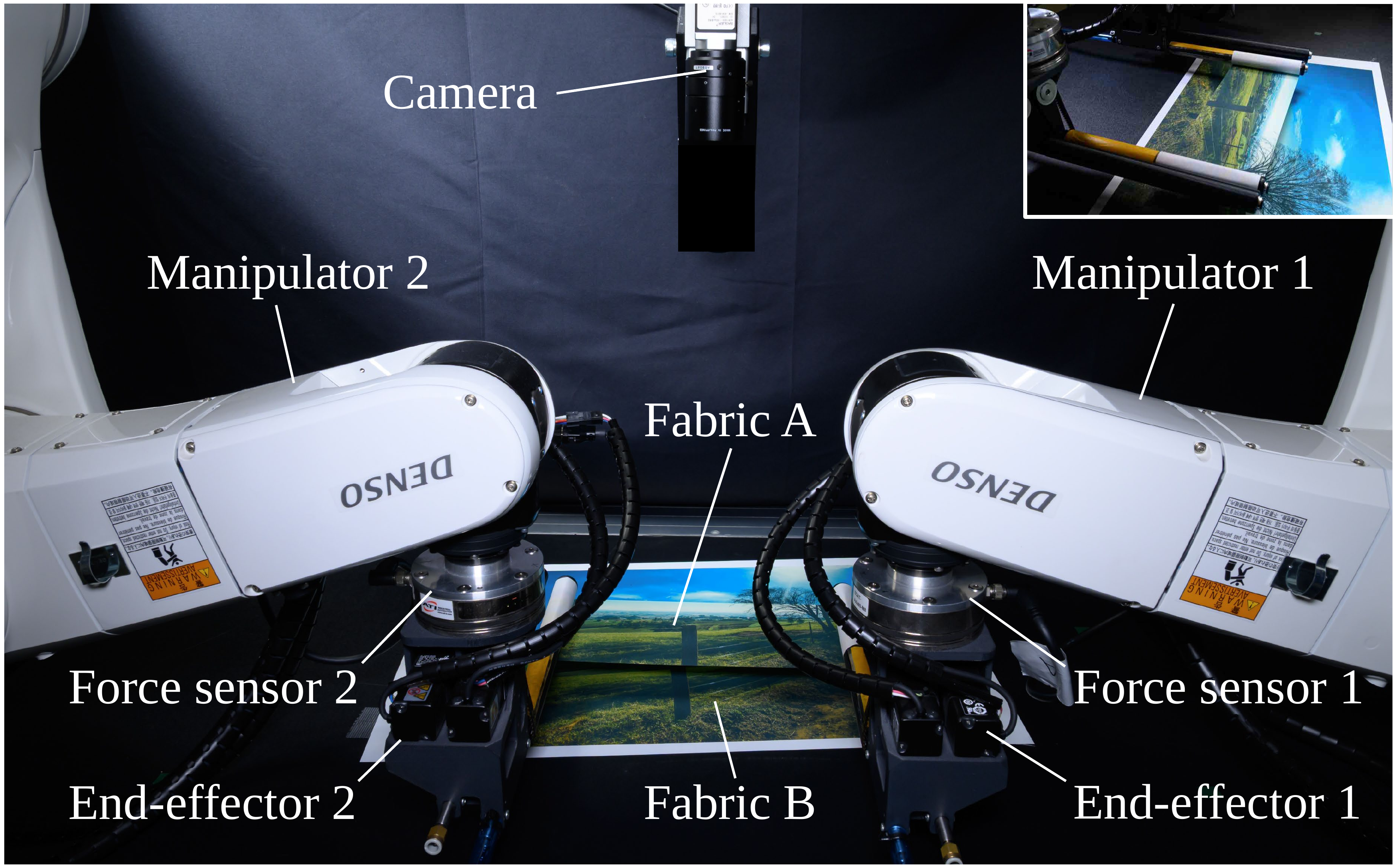}
\caption{Overview of the dual-arm manipulator system used for fabric alignment and placement.\label{SYSTEM}}
\end{figure}

\IEEEPARstart{I}{n} the garment manufacturing process, robotic handling of fabric pieces is one of the technologies being explored to replace human labor. As shown in Fig.~\ref{SYSTEM}, we propose a method for aligning and placing a fabric piece (Fabric~A) on top of another (Fabric~B) so that their surface textures match. This is a common task in garment production, such as aligning the texture of a pocket with that of a shirt.

In our proposed system, both edges of Fabric~A are rolled up and constrained by end-effectors~\cite{kobayashi2025rollup} attached to a dual-arm manipulator. We then introduce a novel motion control scheme based on visual servoing, which enables precise pose control of Fabric~A while maintaining its flatness using the dual-arm manipulator. We assume that Fabric A and Fabric B have the same texture.

In this paper, we propose a novel method that combines Transformer-based visual servoing with dual-arm impedance control, enabling precise alignment and placement of fabric pieces using solely their texture patterns. The Transformer network, trained entirely on synthetic data, achieves zero-shot sim-to-real generalization and robustly aligns a wide range of previously unseen fabric textures. Simultaneously, the dual-arm impedance controller applies internal forces to keep the fabric flat, ensuring that the texture patterns of both fabric pieces are consistent, which is essential for accurate visual servoing.

The contributions of this paper can be summarized as follows:
\begin{enumerate}
  \item We propose a new motion control scheme for a dual-arm manipulator that employs Transformer-driven visual servoing and dual-arm impedance control. This scheme enables precise alignment and placement of a fabric piece on top of another, so that their textures accurately match.
  
  \item We propose a new network architecture for visual servoing based on a pre-trained backbone and a newly proposed Difference Extraction Attention Module (DEAM), which significantly improves the pose difference prediction accuracy.
  
  \item The proposed network is trained entirely on images generated using rendering software, enabling zero-shot application in real-world environments without prior training on specific fabric textures.

  \item The experiments using unseen textures demonstrate that our system can align the fabric piece with an average position error of $0.1~\mathrm{mm}$. We further show through experiments that the system is capable of aligning fabric pieces under various unseen textures and unseen lighting conditions.
\end{enumerate}

\section{RELATED WORK}

Over the years, visual servoing~\cite{F.Chaumette} has evolved significantly, particularly with the advent of deep learning techniques such as convolutional neural networks (CNNs). Recent advances in CNN-based visual servoing~\cite{Saxena, Bateux, Yu, Tokuda, Harish, Tokuda2} have expanded its applications by eliminating the need for manual feature engineering and precise camera calibration. These approaches have demonstrated robust performance in tasks such as connector insertion~\cite{Yu} and assembly~\cite{Tokuda,Tokuda2}, even under challenging conditions such as varying lighting~\cite{Bateux,Tokuda2} and occlusion~\cite{Bateux,Tokuda2}. More recently, CNN-based visual servoing has been extended to deformable object manipulation. Tokuda et al.~\cite{Tokuda3} achieved fabric positioning and flattening using a dual-arm manipulator and a projected pattern to highlight wrinkles. However, the network in their work was trained on real images collected from physical experiments, which limits its generalization to unseen textures and lighting conditions.

In contrast to CNN-based frameworks, several studies have explored non-learning-based visual servoing for deformable object control. Shetab-Bushehri et al.~\cite{shetab2023lattice} proposed a lattice-based tracking and servoing method for controlling elastic object shapes, while Qi et al.~\cite{qi2021contour} introduced a contour-moment-based control strategy with finite-time model estimation for composite rigid–deformable objects. Despite methodological differences, most existing visual servoing approaches for deformable objects share several limitations: (a) reliance on visible edges or contours, restricting applicability under occlusion; (b) limited adaptability to textured deformable objects; and (c) dependence on large-scale real-world datasets.

Beyond visual servoing, recent policy/trajectory learning has demonstrated impressive generalization capabilities in robotic manipulation. Diffusion Policy~\cite{chi2023diffusion} learns visuomotor control through action diffusion, enabling robust imitation learning from demonstration data. Similarly, TraKDis~\cite{chen2024trakdis} adopts a transformer-based knowledge distillation framework for reinforcement learning, achieving effective policy transfer to deformable object tasks such as cloth manipulation. From a planning perspective, Wang et al.~\cite{wang2024efficient} proposed Deformation-Aware RRT*, which incorporates deformation modeling into sampling-based motion planning for planar fabric repositioning. These methods focus primarily on policy learning or trajectory optimization, whereas visual servoing provides real-time, closed-loop control for precise alignment and positioning.

In summary, previous methods have advanced visual servoing and policy/trajectory learning methods for both rigid and deformable objects. However, CNN-based and non-learning visual servoing frameworks still struggle to adapt to different textures and lighting, and rely heavily on real-world data. Inspired by recent progress in policy and trajectory learning that utilizes Transformer-based attention architectures, our approach introduces these capabilities into visual servoing to achieve texture-based alignment using only synthetic data, unifying visual perception and dual-arm impedance control for zero-shot real-world deployment.

\section{COORDINATE SYSTEMS}
\begin{figure}[t]
\centering
\includegraphics[width=85mm]{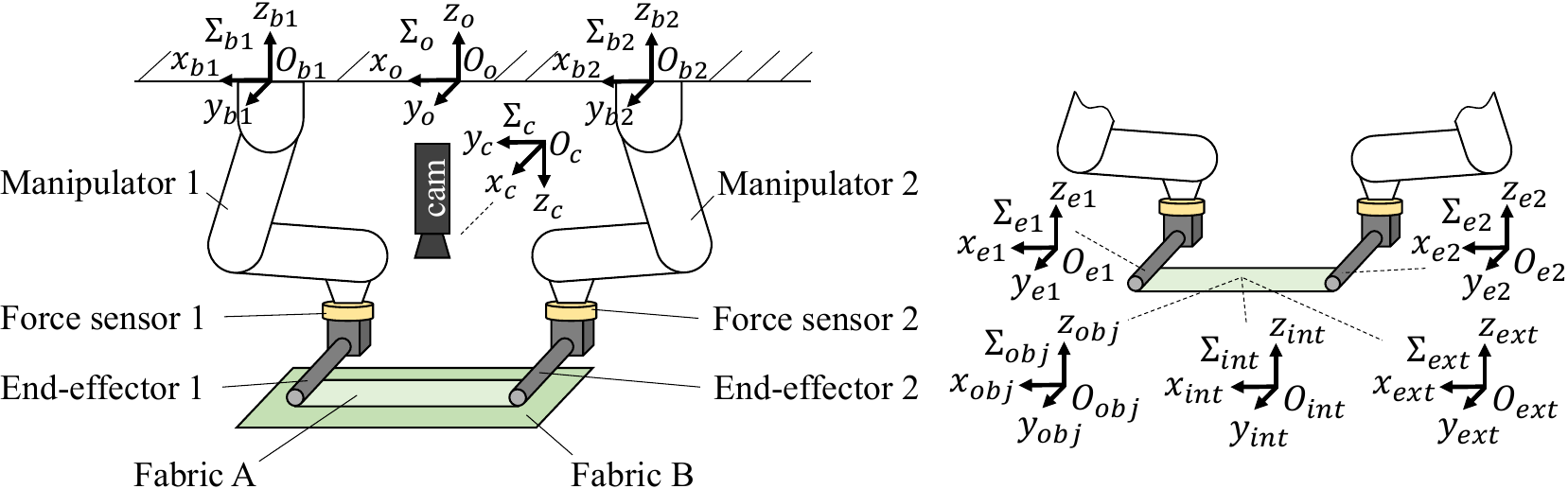}
\caption{Definition of coordinate systems.\label{COORDINATE}}
\end{figure}

The coordinate systems are defined, as shown in Fig.~\ref{COORDINATE}.
The world coordinate system ($\Sigma_{o}=O_{o}{-}x_{o}y_{o}z_{o}$), the bases of the manipulators (Manipulator~1 and Manipulator~2) $\Sigma_{b1}$ and $\Sigma_{b2}$, and the camera coordinate system $\Sigma_{c}$ are related through homogeneous transformations obtained by camera-robot calibration.
Each end-effector coordinate system, $\Sigma_{e1}$ and $\Sigma_{e2}$, is obtained from the corresponding base frame via forward kinematics.

The object coordinate system $\Sigma_{obj}$ represents the pose of Fabric~A and is attached to the midpoint between $\Sigma_{e1}$ and $\Sigma_{e2}$.
Its $z$-axis is aligned with that of $\Sigma_{o}$, and its $x$-axis points from $\Sigma_{e1}$ to $\Sigma_{e2}$.
The internal and external force coordinate systems, $\Sigma_{int}$ and $\Sigma_{ext}$, used for impedance control, are identical to $\Sigma_{obj}$.

\section{PROPOSED METHOD}
\begin{figure*}[t]
    \centering
    \includegraphics[width=166mm]{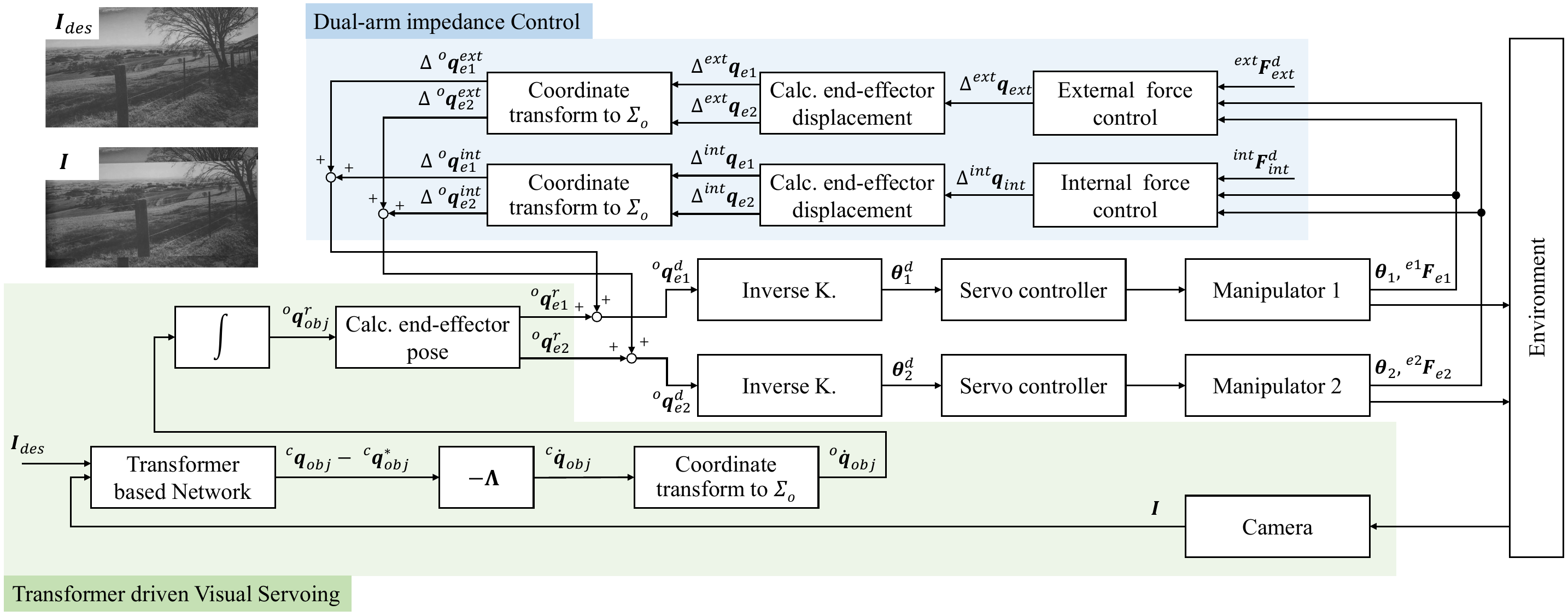}
    \caption{Block diagram of the control system. The proposed control system consists of Transformer-driven visual servoing (in green) and dual-arm impedance control (in blue).\label{BLOCK}}
\end{figure*}

Fig.\ref{BLOCK} shows the block diagram of the proposed control scheme. The Transformer-driven visual servoing (in green) controls the pose of Fabric~A to align its texture with that of Fabric~B, using grayscale images captured by a monocular camera. The dual-arm impedance control (in blue) maintains texture consistency between Fabric~A and Fabric~B by flattening the fabric by controlling the internal force. The details of both the Transformer-driven visual servoing and the dual-arm impedance control are explained in the following subsections.

\subsection{Visual Servoing Control}
\subsubsection{Control Law}
As shown in Fig.~\ref{BLOCK}, the network takes two images as input: the desired image ${\bm I}_{des}$ and the current image ${\bm I}(t)$. The desired image ${\bm I}_{des}$ is an image of Fabric~B captured before each visual servoing execution, while ${\bm I}(t)$ is a grayscale image of Fabric~A and part of Fabric~B captured at time $t$ during visual servoing.


Let ${^{c}{\bm q}_{obj}} = {[^{c}{\bm t}_{obj}^{T}, ^{c}{\bm r}_{obj}^{T}]}^{T}$ be the stack of the three-dimensional position vector and the XYZ Euler rotation vector representing the current Fabric~A pose in the camera coordinate system. Similarly, let ${^{c}{\bm q}_{obj}^{*}}= {[^{c}{\bm t}_{obj}^{*T}, ^{c}{\bm r}_{obj}^{*T}]}^{T}$ denote the desired Fabric~A pose in the camera coordinate system. The output of the network is defined as
\begin{equation}
{^{c}{\bm q}_{obj}(t)}-{^{c}{\bm q}_{obj}^{*}} = f({\bm I}_{des}, {\bm I}(t)),
\label{EQ1}
\end{equation}

\noindent where $f$ is a neural network that maps the image pair $({\bm I}_{des}, {\bm I}(t))$ to the pose difference. Although the subtraction operator is used for notational simplicity, the rotational difference is calculated by converting the poses to rotation matrices or quaternions, instead of subtracting Euler angles directly.

The desired velocity command for visual servoing, ${^{c}\dot{\bm q}_{obj}}$, represented in the camera coordinate system are computed as
\begin{equation}
{^{c}\dot{\bm q}_{obj}(t)} = -{\bm \Lambda} \cdot ({^{c}{\bm q}_{obj}(t)}-{^{c}{\bm q}_{obj}^{*}}),
\label{EQ2}
\end{equation}

\noindent where ${\bm \Lambda} \in \mathbb{R}^{6\times6}$ is a diagonal, positive definite gain matrix defined as
\begin{equation}
{\bm \Lambda} = \mathrm{diag}(\lambda_{tx},\lambda_{ty},\lambda_{tz},\lambda_{rx},\lambda_{ry},\lambda_{rz}),
\label{EQ3}
\end{equation}

\noindent where $\lambda_{tx}$, $\lambda_{ty}$, and $\lambda_{tz}$ are the proportional gains for translational errors along the $x$-, $y$-, and $z$-axes, respectively, and $\lambda_{rx}$, $\lambda_{ry}$, and $\lambda_{rz}$ are the gains for rotational errors. In our experiments, $\lambda_{tz}$, $\lambda_{rx}$, and $\lambda_{ry}$ are defined as zero. $x$-axis translation, $y$-axis translation, and $z$-axis rotation are controlled by visual servoing, while $z$-axis translation, $x$-axis rotation, and $y$-axis rotation are controlled by impedance control.

Then, the desired velocity command ${^{o}\dot{\bm q}_{obj}}$, represented in $\Sigma_o$, is computed by applying a coordinate transformation to ${^{c}\dot{\bm q}_{obj}}$ expressed in $\Sigma_c$. The reference pose ${^{o}{\bm q}_{obj}^{r}}$, which is the desired Fabric~A pose in visual servoing, is calculated as
\begin{equation}
    {^{o}{\bm q}_{obj}^{r}} = {^{o}{\bm q}_{obj}(0)} + \int_{0}^{t + T} {^{o}\dot{\bm q}_{obj}}(\tau) \, d \tau,
  \label{EQ4}
\end{equation}

\noindent where $T$ is the control cycle time of the visual servoing loop. Finally, the reference poses of the end-effectors, ${^{o}{\bm q}_{e1}^{r}}$ and ${^{o}{\bm q}_{e2}^{r}}$, are computed from ${^{o}{\bm q}_{obj}^{r}}$ using the transformations between the object coordinate system and the each end-effector coordinate system.

\subsubsection{Network Architecture}
\begin{figure}[t]
\centering
\includegraphics[width=80mm]{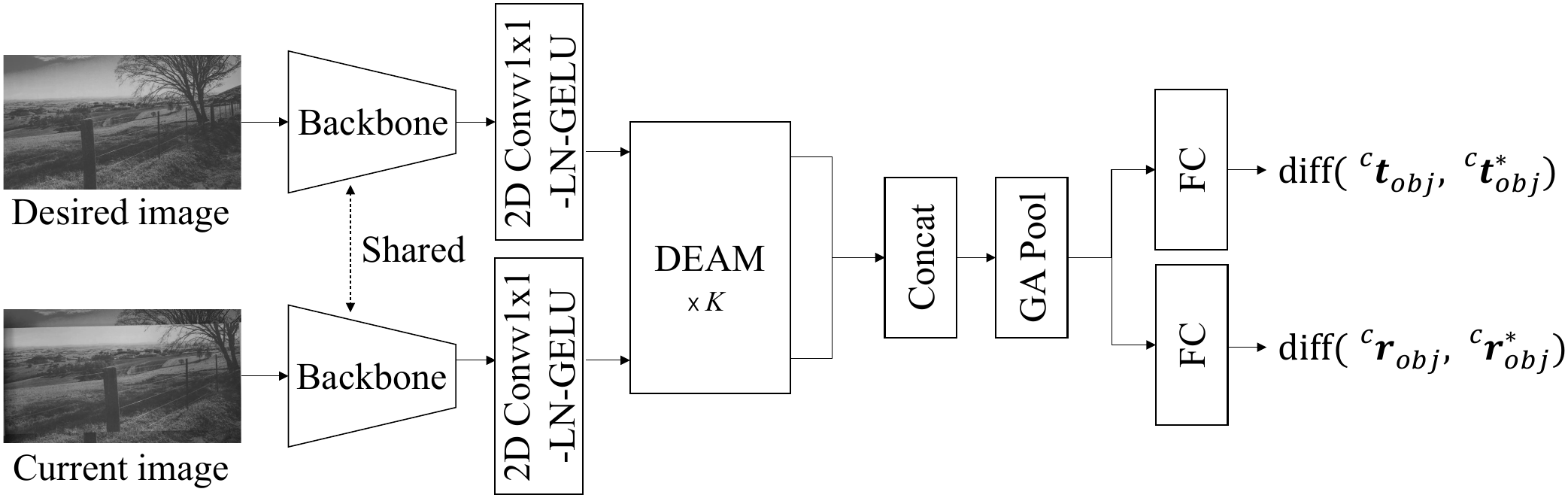}
\caption{Overall architecture of the Transformer-driven visual servoing network. \label{NETWORK1}}
\end{figure}

Conventional CNN-based visual servoing methods typically concatenate or subtract features from Siamese backbones before feeding them into fully connected layers~\cite{Yu, Tokuda2}. This simple concatenation or subtraction-based feature fusion has been widely used for learning pose differences between image pairs due to its simplicity. In contrast, our network adopts a pre-trained Transformer backbone with Difference Extraction Attention Module (DEAM) to handle the fabric alignment task, as shown in Fig.~\ref{NETWORK1}. Because alignment depends on subtle, spatially distributed texture cues rather than distinct geometric landmarks, global feature modeling is essential. The Transformer captures long-range dependencies, while DEAM extracts inter-image feature differences at multiple stages to enhance sensitivity to small texture shifts.

Unlike CNN-based Siamese models that compute differences through simple concatenation or subtraction, our design embeds this comparison process directly into the attention mechanism, enabling adaptive learning of cross-image differences. Compared with attention mechanisms such as SE~\cite{hu2018squeeze}, ECA~\cite{wang2020eca}, and GRN~\cite{woo2023convnext}, DEAM explicitly models cross-image feature interactions rather than applying attention independently within each map.

First, the network processes two input images (current and desired) independently using shared-weight backbones. The features from each backbone are first projected into a lower-dimensional embedding space via a \texttt{1×1} convolutional layer (kernel size of one), followed by Layer Normalization and a GELU activation function (2D Conv\texttt{1×1}–LN–GELU). The features are then passed into the $K$ layers of Difference Extraction Attention Module (DEAM) to progressively extract feature differences. The two feature maps output from the final DEAM layer are concatenated, globally average pooled (GA Pool), and then passed into fully connected layers (FC) to predict the 6-DoF pose difference (translation and rotation) of Fabric~A.

\begin{figure*}[t] 
    \centering
    \includegraphics[width=166mm]{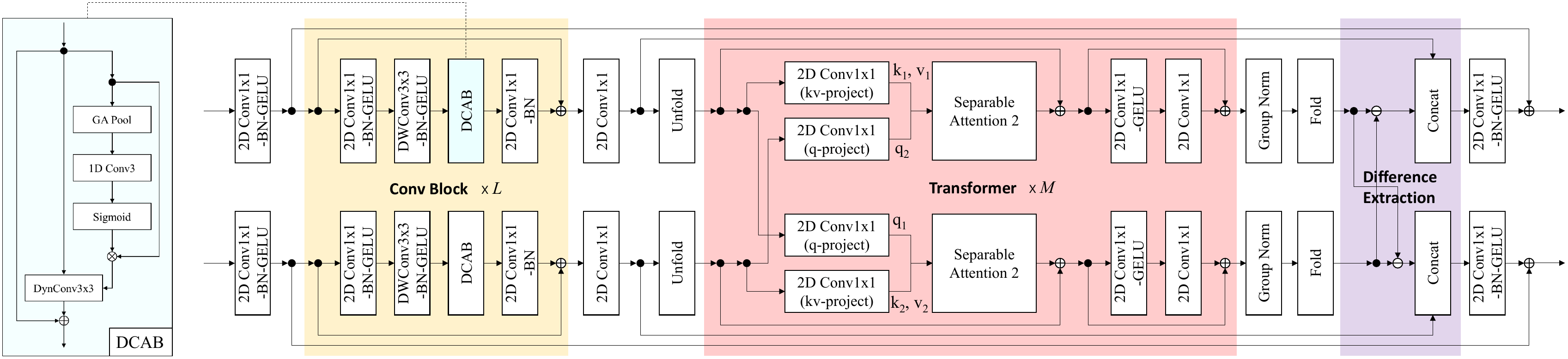}
    \caption{Architecture of the Difference Extraction Attention Module (DEAM), which combines convolutional blocks, Transformer blocks, and a difference extraction blocks to enhance prediction accuracy. The convolutional block includes the proposed Dynamic Convolution by Attention Blocks (DCAB). \label{NETWORK2}}
\end{figure*}

As shown in Fig.\ref{NETWORK2}, each DEAM comprises three components: (1) Convolutional Blocks, (2) Transformer Blocks, and (3) a Difference Extraction Block. DEAM begins with a stack of Convolutional Blocks that employ the channel expansion–compression strategy from EfficientNet\cite{tan2019efficientnet}. Each block consists of a \texttt{1×1} convolutional layer, followed by Batch Normalization and a GELU activation function (2D Conv\texttt{1×1}–BN–GELU), and then a depthwise \texttt{3×3} convolutional layer with Batch Normalization and GELU activation (DW Conv\texttt{3×3}–BN–GELU). The resulting features are subsequently fed into a proposed module named Dynamic Convolution by Attention Blocks (DCAB), as illustrated in Fig.~\ref{NETWORK2}.

DCAB is built upon a dynamic convolutional layer~\cite{Chen} with a \texttt{3×3} kernel (DynConv\texttt{3×3}), whose convolution kernels are dynamically generated from input-dependent attention scores computed using Efficient Channel Attention (ECA)~\cite{wang2020eca}. The output of DCAB is fed into the final layer of the Convolutional Block: a 2D Conv\texttt{1×1}–BN layer that projects the features back to their original channel dimension. After $L$ layers of the Convolutional Block, the extracted features are projected into a lower-dimensional space by a 2D Conv\texttt{1×1} layer, unfolded~\cite{Mehta1}, and then passed into Transformer Blocks.

The Transformer Block is a cross-attention mechanism~\cite{Vaswani} that applies a separable attention mechanism~\cite{Mehta2}. Separable attention improves computational efficiency by decoupling the attention operation into spatial and channel-wise components. The Transformer mechanism takes "key" and "value" features from one image and "query" features from the other, processes them through separable attention, and then through a feed-forward network consisting of 2D Conv\texttt{1×1}–GELU and 2D Conv\texttt{1×1} layers. The output features are then folded back to their original spatial dimensions. The Transformer Block is repeated $M$ layers. In our design, the cross-attention in the desired image branch uses query features from the current image and key/value features from the desired image. The query features indicate the regions to be aligned, while the key/value features provide the reference appearance and context. This allows the network to find correspondences between the two feature spaces. Conversely, the cross-attention in the current image branch learns complementary relationships in the opposite direction.

To enhance sensitivity to feature-level differences, the element-wise difference between the two streams is concatenated with each of the original feature maps. The resulting features are passed through a 2D Conv\texttt{1×1}–BN–GELU layer to project them back to their original channel dimension. 

\subsubsection{Training Dataset}
The training dataset is generated using the rendering software Blender. The end-effectors, Fabric~A, Fabric~B, the camera, and LED light sources are all simulated within the environment. The physics engine in Blender is used to simulate fabric deformation, and image rendering is performed using the Cycles renderer. During each rendering session, the poses of the fabric pieces and the camera, as well as the intensity and positions of the light sources, are randomly varied within predefined ranges. 

\begin{figure}[t]
\centering
\includegraphics[width=70mm]{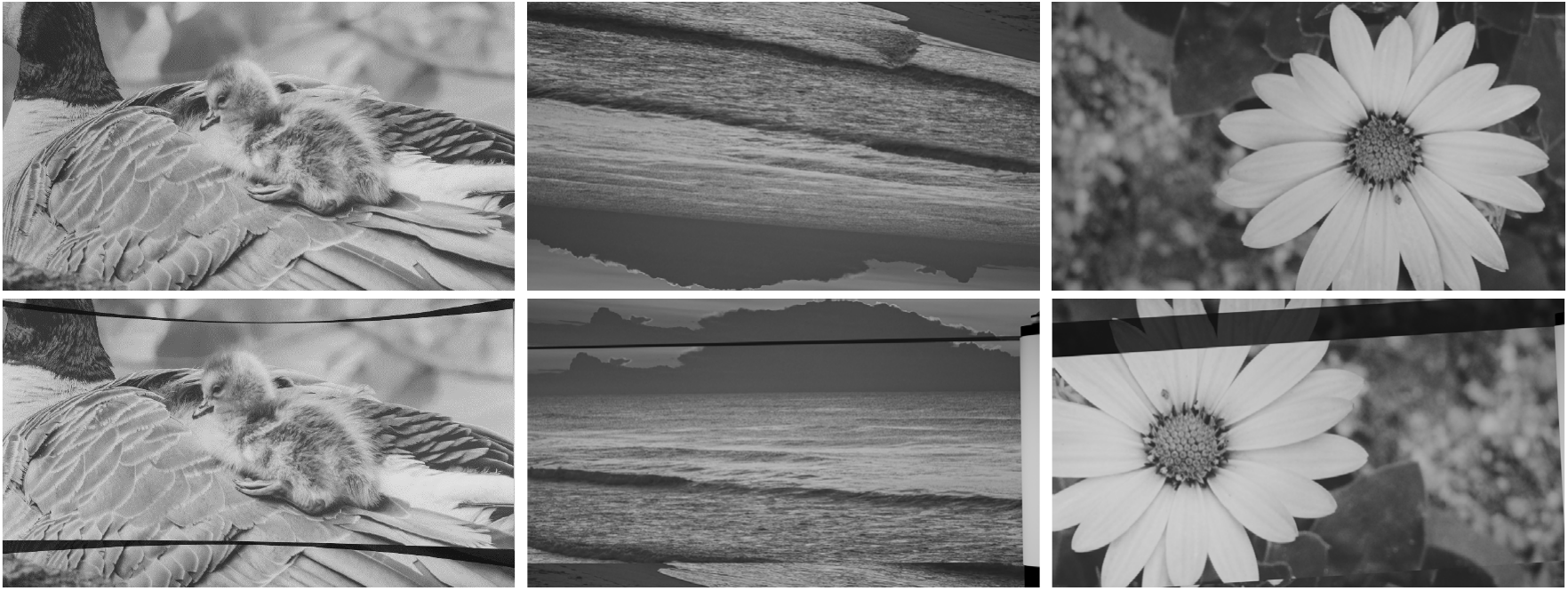}
\caption{Examples of synthetic image pairs used in training. Top row: desired images. Bottom row: corresponding current images. \label{DATASET}}
\end{figure}

To generate diverse fabric deformations, the relative position between the two end-effectors grasping the fabric is also randomly changed within a specified range. In addition, the number of light sources is randomly varied. The textures of the fabric pieces are randomly assigned using images obtained via the Pexels API. Examples of the rendered images are shown in Fig.~\ref{DATASET}, where the first row contains the desired images and the second row shows the current images.

\subsection{Dual-Arm Impedance Control}
To flatten the fabric piece during the visual servoing, dual-arm impedance control is utilized based on coordinated motion control proposed by Kosuge et al.\cite{kosuge1997coordinated,Kosuge4} to control both the external and internal forces applied to the end-effectors and Fabric~A.

\subsubsection{External Force Control}
Let ${^{ext}{\bm F}_{ext}} = {[^{ext}{\bm f}_{ext}^{T}, ^{ext}{\bm \eta}_{ext}^{T}]}^{T} \in \mathbb{R}^6$ denote the equivalent force and torque at the origin of the external force coordinate system, applied by the two end-effectors. ${^{ext}{\bm F}_{ext}}$ is computed from the forces and torques applied to the two end-effectors, ${^{e1}{\bm F}_{e1}} \in \mathbb{R}^6$ and ${^{e2}{\bm F}_{e2}} \in \mathbb{R}^6$. For details, please refer to~\cite{Kosuge4}.

The external force is controlled according to the dynamics of a virtual mechanism based on the impedance model, as
\begin{multline}
{\bm M}_{ext} \Delta ^{ext}\ddot{\bm q}_{ext} + {\bm D}_{ext} \Delta ^{ext}\dot{\bm q}_{ext} + {\bm K}_{ext} \Delta ^{ext}{\bm q}_{ext} \\
= {\bm S}_{ext}({^{ext}{\bm F}_{ext}}-{^{ext}{\bm F}_{ext}^{d}}),
\end{multline}
\noindent where ${\bm M}_{ext} \in \mathbb{R}^{6 \times 6}$, ${\bm D}_{ext} \in \mathbb{R}^{6 \times 6}$, and ${\bm K}_{ext} \in \mathbb{R}^{6 \times 6}$ are the inertia, damping, and stiffness matrices for external force control, respectively. $\Delta {^{ext}{\bm q}_{ext}}$ denotes the displacement from the reference pose of Fabric~A, ${^{o}{\bm q}_{obj}^{r}}$, which is computed by the visual servoing control described in the previous subsection. ${^{ext}{\bm F}_{ext}^{d}} \in \mathbb{R}^{6}$ is the desired external force expressed in the external force coordinate system. The selection matrix ${\bm S}_{ext} = \mathrm{diag}(0, 0, 1, 1, 1, 0) \in \mathbb{R}^{6 \times 6}$ determines which axes are controlled by external force control.

The displacement of each end-effector, $\Delta {^{o}{\bm q}_{e1}}$ and $\Delta {^{o}{\bm q}_{e2}}$, is calculated from $\Delta {^{ext}{\bm q}_{ext}}$ using the transformation matrices between the external force coordinate system and the two end-effector coordinate systems. 

\subsubsection{Internal Force Control}
Let us define ${^{int}{\bm F}_{int}} = {[^{int}{\bm f}_{int}^{T}, ^{int}{\bm \eta}_{int}^{T}]}^{T} \in \mathbb{R}^6$ as the internal force applied by the two end-effectors of Manipulator~1 and Manipulator~2. ${^{int}{\bm F}_{int}}$ is computed based on the method described in~\cite{Kosuge4}.

The internal force is controlled based on the model as
\begin{multline}
{\bm M}_{int} \Delta ^{int}\ddot{\bm q}_{int} + {\bm D}_{int} \Delta ^{int}\dot{\bm q}_{int} =\\
{\bm S}_{int}({^{int}{\bm F}_{int}} - {^{int}{\bm F}_{int}^{d}}),
\end{multline}

\noindent where ${\bm M}_{int} \in \mathbb{R}^{6 \times 6}$ and ${\bm D}_{int} \in \mathbb{R}^{6 \times 6}$ are the inertia and damping matrices for internal force control, respectively.
$\Delta {^{int}{\bm q}_{int}}$ denotes the displacement from the initial relative pose of the two end-effectors with respect to the internal force coordinate system.
${^{int}{\bm F}_{int}^{d}} \in \mathbb{R}^{6}$ is the desired internal force expressed in the internal force coordinate system.
The selection matrix ${\bm S}_{int} = \mathrm{diag}(1, 0, 0, 0, 0, 0) \in \mathbb{R}^{6 \times 6}$ determines which axes are controlled by internal force control.

The displacements of each end-effector, $\Delta {^{o}{\bm q}_{e1}}$ and $\Delta {^{o}{\bm q}_{e2}}$ are computed from $\Delta {^{int}{\bm q}_{int}}$ using the transformation matrices between the internal force coordinate system and the end-effector coordinate systems. Finally, the desired pose of the end-effectors of Manipulator~1 and Manipulator~2 are computed as
\begin{equation}
{^{o}{\bm q}_{e1}^{d}} = {^{o}{\bm q}_{e1}^{r}} + \Delta {^{o}{\bm q}_{e1}^{ext}} + \Delta {^{o}{\bm q}_{e1}^{int}},
\end{equation}
\begin{equation}
{^{o}{\bm q}_{e2}^{d}} = {^{o}{\bm q}_{e2}^{r}} + \Delta {^{o}{\bm q}_{e2}^{ext}} + \Delta {^{o}{\bm q}_{e2}^{int}}.
\end{equation}

\noindent The desired joint angles of each manipulator commanded to each robot servo controller, ${\bm \theta}_{1}^{d} \in \mathbb{R}^{6}$ and ${\bm \theta}_{2}^{d}\in \mathbb{R}^{6}$, are obtained by solving the inverse kinematics based on the desired end-effector poses.

\section{EXPERIMENTAL RESULTS}
\subsection{Network Comparison} 
For the backbone network, we adopt the Vision Transformer (ViT)~\cite{dosovitskiy2020image} with a patch size of 32, considering both prediction accuracy and computational cost. The backbone networks are pre-trained on ImageNet. The entire network, including the two backbones and the DEAM ($K=1$, $L=1$, $M=1$), is trained for 100 epochs using the AdamW~\cite{loshchilov2017decoupled} optimizer, with a batch size of 64 (8 samples per GPU $\times$ 8 GPUs).

We use a total of 100{,}000 samples for training (with 10\% held out for validation) and 58{,}000 samples for testing. The input size of both the current and desired images is $960 \times 540~\mathrm{pixels}$ (width $\times$ height). The learning rate is linearly increased from \(1 \times 10^{-6}\) to \(1 \times 10^{-4}\) over the first 5 epochs, and then decayed to \(1 \times 10^{-5}\) using a cosine annealing schedule~\cite{Polyak}.
The loss function is defined as
\begin{multline}    
    E = \alpha \left\| {\widehat{({^{c}{\bm t}_{obj}}-{^{c}{\bm t}_{obj}^{*}})}} - {({^{c}{\bm t}_{obj}}-{^{c}{\bm t}_{obj}^{*}})} \right\|_2 \\
    + \beta \left\| {\widehat{({^{c}{\bm r}_{obj}}-{^{c}{\bm r}_{obj}^{*}}})} - {({^{c}{\bm r}_{obj}}-{^{c}{\bm r}_{obj}^{*}})} \right\|_2,
\end{multline}

\noindent
where $\widehat{({^{c}{\bm t}_{obj}}-{^{c}{\bm t}_{obj}^{*}})}$ and $\widehat{({^{c}{\bm r}_{obj}}-{^{c}{\bm r}_{obj}^{*}})}$ denote the predicted translational and rotational difference, respectively.
$({^{c}{\bm t}_{obj}}-{^{c}{\bm t}_{obj}^{*}})$ and $({^{c}{\bm r}_{obj}}-{^{c}{\bm r}_{obj}^{*}})$ denote the ground-truth of translational and rotational difference, respectively. The coefficients $\alpha = 1.0$ and $\beta = 1.0$ are used to weight the translation and rotation losses. Note that both the input images and output vectors are normalized to the range $[0.0, 1.0]$ using min–max normalization based on their respective minimum and maximum values. In our dataset, the fabric piece is randomly moved within a translation range of $\pm 20~\mathrm{mm}$ and a rotation range of $\pm 10^\circ$ around each axis.

Table~\ref{tab:comparison} shows a comparison between the proposed network and several other architectures.
The \textsc{Concat}~\cite{Yu} method employs a Siamese architecture with two shared-weight ViT~\cite{dosovitskiy2020image} backbones whose extracted features are concatenated and fed into fully connected layers. The proposed method, \textsc{DEAM (DCAB)}, achieves the highest accuracy, significantly reducing the loss $\mathit{E}$ (defined in eq.~(9)) compared to \textsc{Concat}. 
A Wilcoxon signed-rank test on the paired errors from the same 58,000 test samples confirmed a significant difference between the two models ($W = 3.04\times10^{8}, p < 0.001, r = 0.55$), indicating that DEAM (DCAB) outperformed \textsc{Concat} with a large effect size.
This demonstrates the effectiveness of DEAM with DCAB, which enhances pose difference prediction accuracy by explicitly capturing feature differences through a combination of convolutional, transformer, and difference extraction blocks. When the difference extraction block is removed, as in \textsc{DEAM (DCAB, w/o Diff. Ext. Block)}, performance degrades notably, highlighting its essential role.

We also evaluated DEAM variants in which the DCAB module is replaced with existing attention mechanisms: \textsc{DEAM (SE~\cite{hu2018squeeze})}, \textsc{DEAM (ECA~\cite{wang2020eca})}, and \textsc{DEAM (GRN~\cite{woo2023convnext})}. Among them, \textsc{DEAM (SE)} and \textsc{DEAM (GRN)} show relatively strong performance, but none outperform \textsc{DEAM (DCAB)}. Although all DEAM variants incur slightly higher GFLOPs than \textsc{Concat}, their inference latency remains below 0.017 seconds per image with a batch size of 1.

Finally, as a supplementary comparison, we applied \textsc{SE}~\cite{hu2018squeeze}, \textsc{ECA}~\cite{wang2020eca}, and \textsc{GRN}~\cite{woo2023convnext} independently to each of the two feature maps extracted from the shared backbone, before concatenation and fully connected layers. While these attention modules are typically designed to be embedded within backbone networks, we include them here to provide additional points of reference. Their performance is notably lower than that of the DEAM-based methods, highlighting the advantage of DEAM's design, which explicitly extracts image feature differences for improved prediction accuracy.

\begin{table*}[t] 
\centering
\caption{Performance Comparison of Different Network Architectures}
\begin{threeparttable}
\begin{tabular}{lccccc}
\hline
\textbf{Method} & Loss $\mathit{E} \; [\times 10^{-3}]\downarrow$ & Trans / Rot RMSE $\downarrow$ & Std$\; [\times 10^{-3}]\downarrow$ & GFLOPs $\downarrow$ & Latency [sec/image] $\downarrow$ \\
\hline
\textsc{Concat~\cite{Yu}} & 8.067 & 6.650 / 1.879 & 2.411 & {\bf 78.272} & {\bf 0.0157} \\
\textsc{DEAM (DCAB)} & {\bf 3.831} & {\bf1.654} / {\bf 1.341} &  {\bf1.410} & 81.865 & 0.0167 \\
\textsc{DEAM (DCAB, w/o Diff. Ext. Block)} & 4.720 & 1.975 / 1.664  & 1.677 & 81.769 & 0.0167 \\
\textsc{DEAM (SE)} & 4.073 &  1.786 / 1.419  & 1.482 & 81.850 & 0.0165 \\
\textsc{DEAM (ECA)} & 4.460 & 1.923 / 1.560 &  1.601 & 81.845 & 0.0165 \\
\textsc{DEAM (GRN)} & 4.049 & 1.741 / 1.417 & 1.469 & 81.843 & 0.0165 \\
\textsc{SE~\cite{hu2018squeeze}} & 7.968 & 6.658 / 1.843 & 2.382 & 78.273 & 0.0158 \\
\textsc{ECA~\cite{wang2020eca}} & 8.006 & 6.544 / 1.880 & 2.407 & 78.273 & 0.0158 \\
\textsc{GRN~\cite{woo2023convnext}} & 8.062 & 6.632 / 1.881 & 2.411 & 78.272 & 0.0158 \\

\hline
\end{tabular}
\begin{tablenotes}
\footnotesize
\item[$\dagger$] $\mathit{E}$ represents the loss computed using normalized network outputs, defined in Eq.~(9), evaluated on 58{,}000 test samples. Translation and Rotation RMSE values are denormalized and reported in millimeters [mm] and degrees [deg], respectively. Std denotes the standard deviation of $\mathit{E}$ over the test set. Inference latency is measured per image with batch size = 1.
\end{tablenotes}
\end{threeparttable}
\label{tab:comparison}
\end{table*}

\subsection{Experimental Results of Visual Servoing }
The system used for the experiments is shown in Fig.~\ref{SYSTEM}. It consists of two 6-DOF manipulators (Denso VS-068), each equipped with a force sensor (ATI Axia80) mounted on the wrist, and a camera (Basler acA1920-155um) that captures grayscale images at a resolution of $960 \times 540~\mathrm{pixels}$ and a frame rate of $60~\mathrm{fps}$. The spatial resolution of the camera is $0.0249~\mathrm{mm/pixel}$. FabricA is a rectangular piece measuring $100 \times 400~\mathrm{mm}$, and FabricB measures $297 \times 420~\mathrm{mm}$. Each end-effector is a custom-built ``Roll-Up"~\cite{kobayashi2025rollup}, which rolls up the fabric edge around a cylindrical surface to securely grasp it without slippage. The end-effectors were designed and fabricated in-house based on our previous work.

Five different textures (Texture~1 to 5), which are not included in the training dataset, are used in the experiments. To ensure the alignment accuracy in real-world experiments, the proposed \textsc{DEAM (DCAB)} ($K=5$, $L=\{2,2,3,3,3\}$, $M=\{2,2,3,3,4\}$) is trained on an extended dataset comprising 200,000 samples for 300 epochs. When running on the robot-control PC equipped with an NVIDIA RTX~4090 GPU, the network exhibited a latency of approximately 34.5~ms during visual servoing. The visual servoing gains are tuned empirically.

The experiments are carried out according to the following procedure:
(1)~Fabric~A and Fabric~B are randomly placed on a flat table.
(2)~An image of Fabric~B is captured and stored as the desired image.
(3)~Fabric~A is detected (the detection algorithm is omitted due to page limitations)
and grasped by the robots' end-effectors.
(4)~The grasped Fabric~A is moved to a fixed pose above Fabric~B.
(5)~The proposed control scheme using \textsc{DEAM (DCAB)} is initiated.

\subsubsection{Visual Servoing of Unseen Textures}
\begin{figure}[t]
    \centering
    \subfigure[Initial image]{%
        \includegraphics[width=27mm]{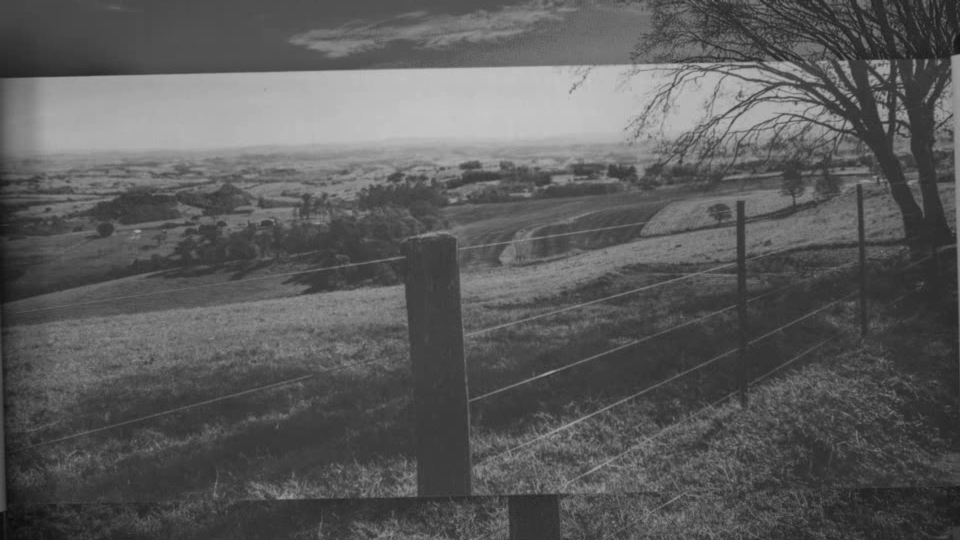}%
        \label{FIG_TEXTURE1_RESULTS1}%
    }\hspace{1mm}%
    \subfigure[Final image]{%
        \includegraphics[width=27mm]{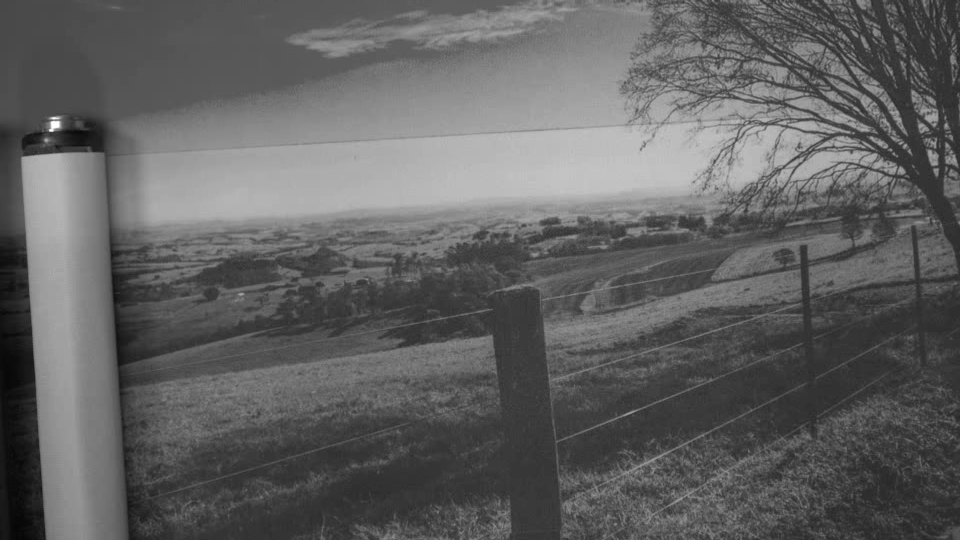}%
        \label{FIG_TEXTURE1_RESULTS2}%
    }\hspace{1mm}%
    \subfigure[Initial error image]{%
        \includegraphics[width=27mm]{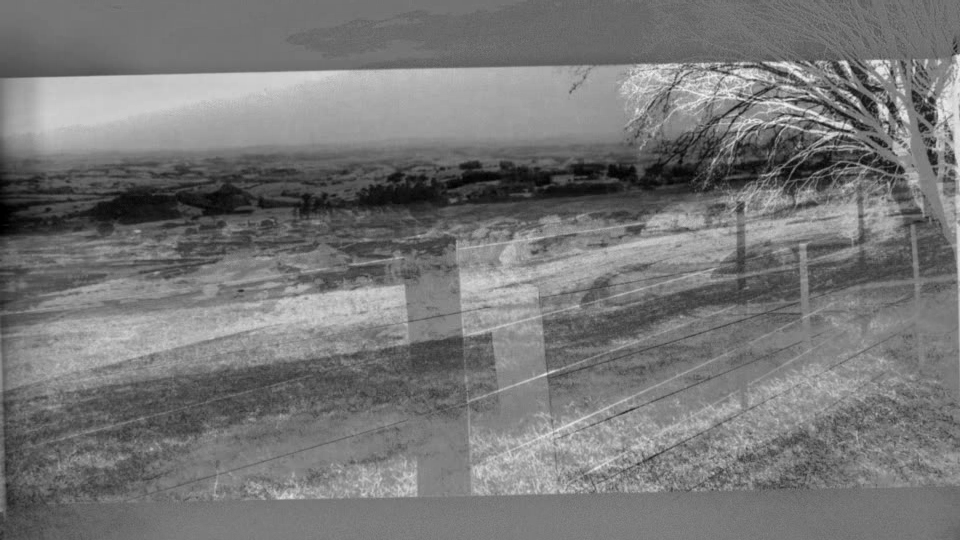}%
        \label{FIG_TEXTURE1_RESULTS3}%
    }
    \subfigure[Final error image]{%
        \includegraphics[width=27mm]{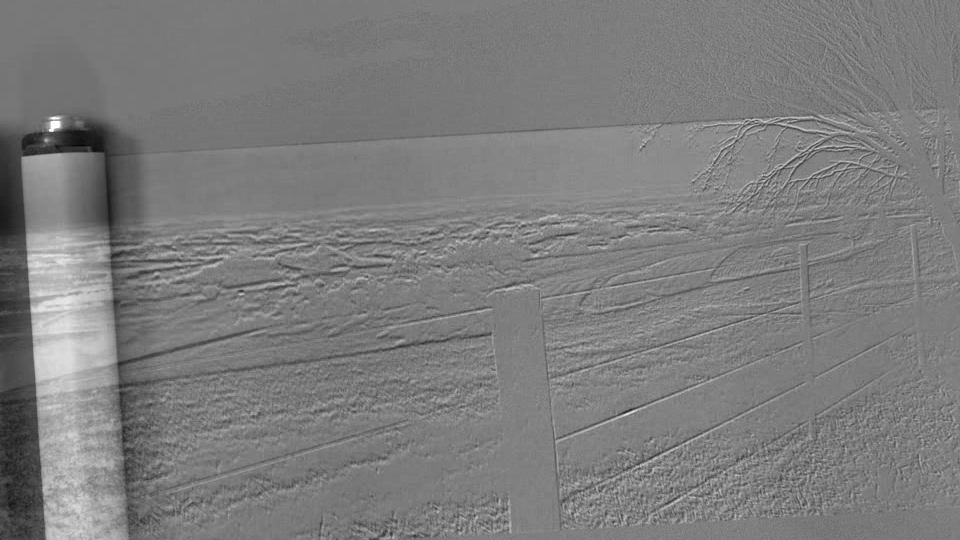}%
        \label{FIG_TEXTURE1_RESULTS4}%
    }\hspace{1mm}%
    \subfigure[Desired image]{%
        \includegraphics[width=27mm]{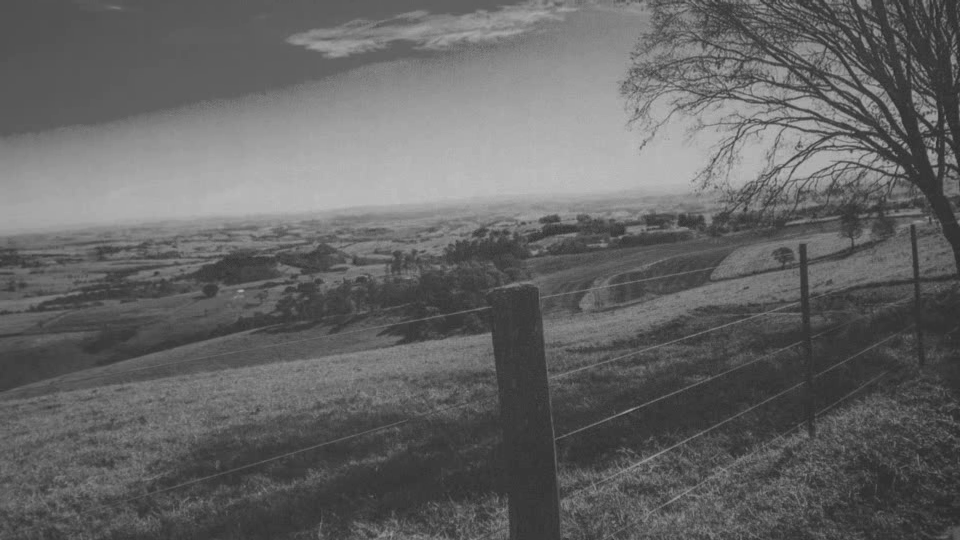}%
        \label{FIG_TEXTURE1_RESULTS5}%
    }\hspace{1mm}%
    \subfigure[SSD]{%
        \raisebox{-1mm}{\includegraphics[width=27mm]{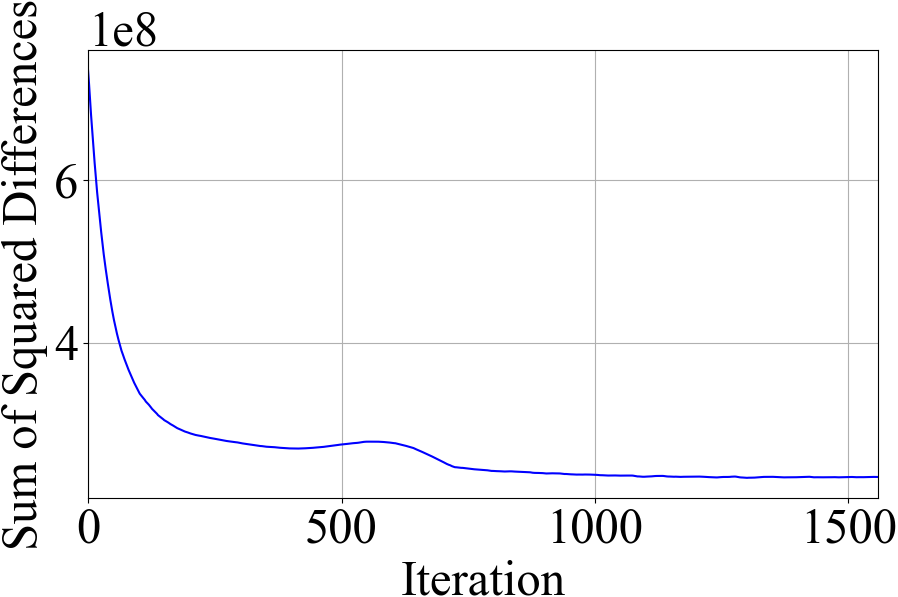}}%
        \label{FIG_TEXTURE1_RESULTS6}%
    }
    \subfigure[Translational displacement]{
        \includegraphics[width=40mm]{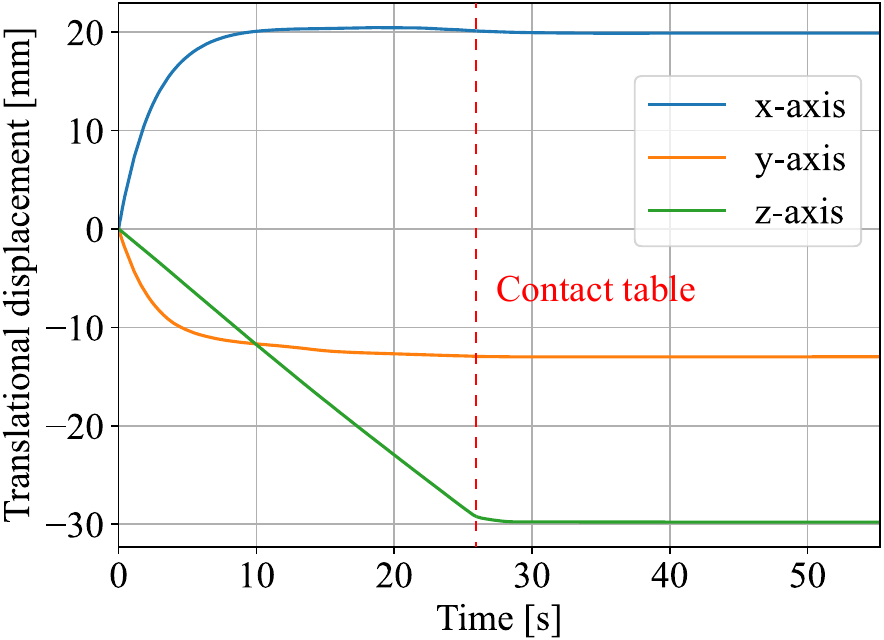}
        \label{FIG_TEXTURE1_RESULTS7}
    }
    \subfigure[Rotational displacement]{
        \includegraphics[width=40mm]{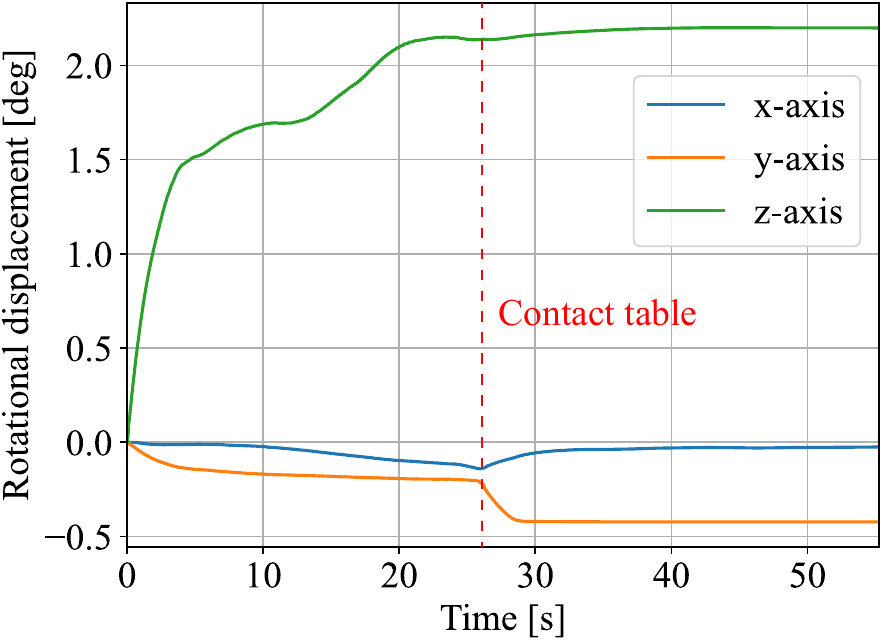}
        \label{FIG_TEXTURE1_RESULTS8}
    }

    \caption{Visual servoing result for Texture~1 using the proposed scheme. 
    (a) Initial image.
    (b) Final image after alignment. 
    (c,d) Error image (${\bm I}-{\bm I}_{des}$) before and after visual servoing. 
    (e) Desired image.
    (f) Sum of squared differences (SSD) between ${\bm I}$ and ${\bm I}_{des}$ over time.
    (g,h) Translational and rotational displacement of the object coordinate system from the initial pose.}
    \label{FIG_TEXTURE1_RESULTS}
\end{figure}

Fig.~\ref{FIG_TEXTURE1_RESULTS} shows an example result using Texture~1. Texture~1 depicts an open field with a distinctive fence at the center and a tree in the upper right, providing strong visual cues. The proposed control scheme successfully aligns and places Fabric~A onto Fabric~B in all 30 trials. After visual servoing, the image error (${\bm I} - {\bm I}_{des}$) is significantly reduced compared to the initial state, as shown in Fig.~\ref{FIG_TEXTURE1_RESULTS}~\subref{FIG_TEXTURE1_RESULTS3} and~\subref{FIG_TEXTURE1_RESULTS4}. The sum of squared differences (SSD) between the desired and current images decreases through visual servoing as shown in Fig.~\ref{FIG_TEXTURE1_RESULTS}~\subref{FIG_TEXTURE1_RESULTS6}. Despite occlusion caused by the end-effectors, the proposed method maintains accurate alignment, demonstrating robustness to occlusion. As shown in Fig.~\ref{FIG_TEXTURE1_RESULTS}~\subref{FIG_TEXTURE1_RESULTS7} and~\subref{FIG_TEXTURE1_RESULTS8}, the Fabric~A pose converges through visual servoing. The figures also show that the end-effectors make contact with the table at around 23 seconds. 

To quantitatively evaluate the positioning accuracy, we compute the average distance between corresponding feature points matched between the final and desired images over 30 visual servoing trials. Note that, due to the difficulty of measuring the actual poses of Fabric A and B, we instead evaluate the average distance between corresponding feature points. Feature points are extracted using SIFT, and mismatches or irrelevant correspondences are manually removed. Using the remaining 7,478 matched points, we calculate the average Euclidean distance between the corresponding points based on camera parameters. The overall average positioning error is $0.106~\mathrm{mm}$, with a standard deviation of $0.043~\mathrm{mm}$ and a maximum error of $0.450~\mathrm{mm}$. 

\begin{figure}[t]
    \centering
    \subfigure[Initial error image]{
        \includegraphics[width=35mm]{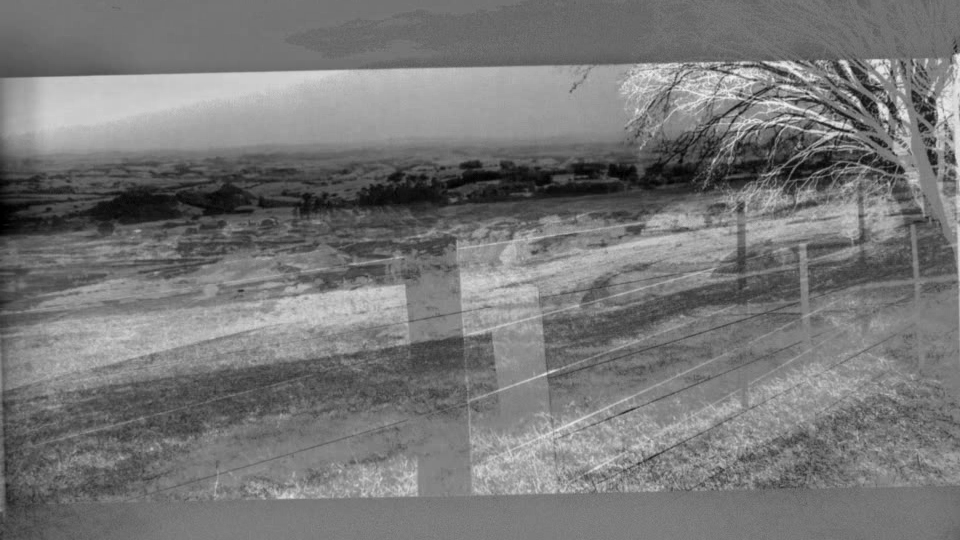}
    }
    \subfigure[Final error image]{
        \includegraphics[width=35mm]{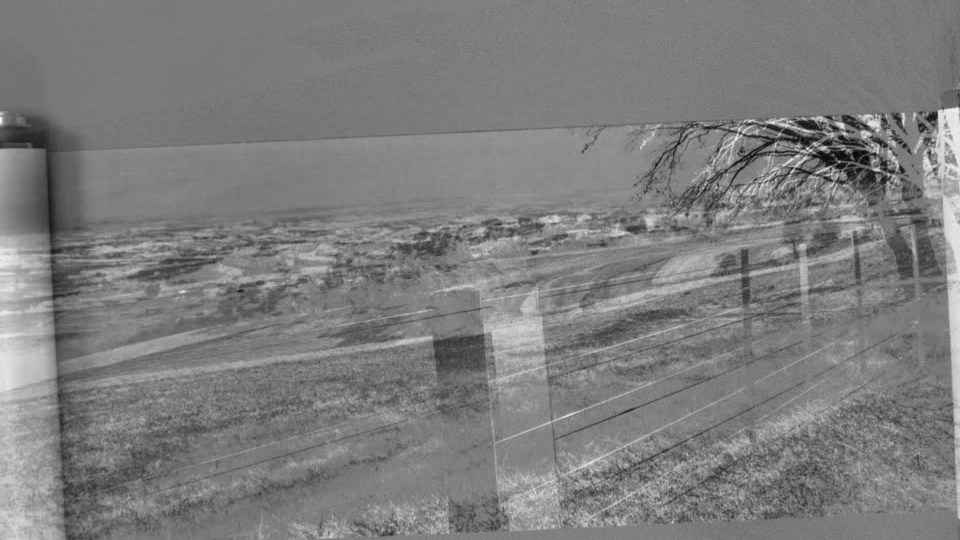}
    }\\
    \caption{Result of direct visual servoing for Texture~1. (a, b) Error images before and after visual servoing.} 
    \label{FIG_PVS_COMPARISON}
\end{figure}

Fig.~\ref{FIG_PVS_COMPARISON} presents the result of direct visual servoing~\cite{C.Collewet} under the same conditions. Unlike the proposed method, direct visual servoing often fails to converge and sometimes even diverges, resulting in low positioning accuracy. However, it is worth noting that when the initial pose error is small (within approximately $3~\mathrm{mm}$ and $0.5^\circ$), direct visual servoing can achieve remarkably good performance. The failure under larger errors is mainly due to two factors. First, the image-based cost function is highly nonlinear with respect to the pose. When the initial error is large, the optimization can easily get trapped in local minima. Second, as the robot approaches the desired pose, the end-effectors occlude the fabric piece in the camera view, causing an appearance mismatch between the current and desired images. This occlusion severely affects the convergence of direct visual servoing.

\begin{figure}[t]
\centering
\includegraphics[width=85mm]{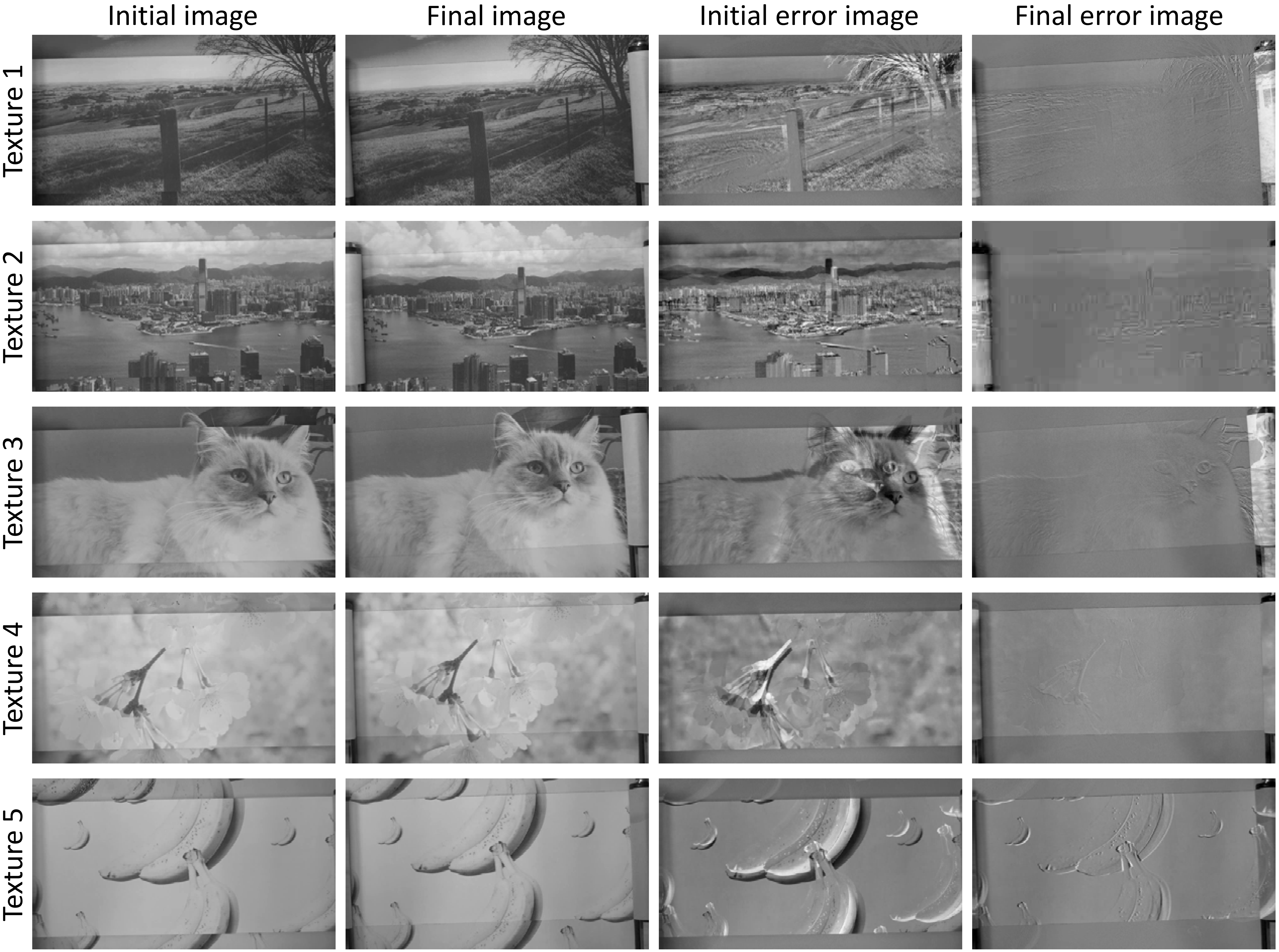}
\caption{Qualitative results of the proposed visual servoing method across all five textures (Texture~1 to 5).\label{FIG_TEXTURE_ALL_RESULTS}}
\end{figure}

Fig.~\ref{FIG_TEXTURE_ALL_RESULTS} summarizes the qualitative results of the proposed visual servoing using Texture~1 to 5. The proposed control scheme successfully aligns Fabric~A with Fabric~B across all textures, demonstrating strong generalization to previously unseen textures. Each texture presents unique challenges. Texture~2 is a landscape image captured by the author in Hong Kong, characterized by dense, detailed patterns that provide rich visual features. Texture~3 contains a distinctive object, a cat, located near the center of the image. Texture~4 features a sharply focused cherry blossom at the center, while the surrounding blurred blossoms result in a low-contrast appearance. Texture~5 is the most challenging, as it contains sparsely distributed patterns of bananas with low saliency and exhibits a low-contrast appearance, providing only weak visual cues for feature extraction. Despite these limitations, the proposed system achieves a reasonable level of alignment even for Texture~5. These results highlight the robustness and generalization capability of the proposed network, demonstrating successful alignment even with previously unseen and visually challenging textures.


\subsubsection{Visual Servoing Under Various Lightning Conditions}
\begin{figure}[t]
\centering
\includegraphics[width=85mm]{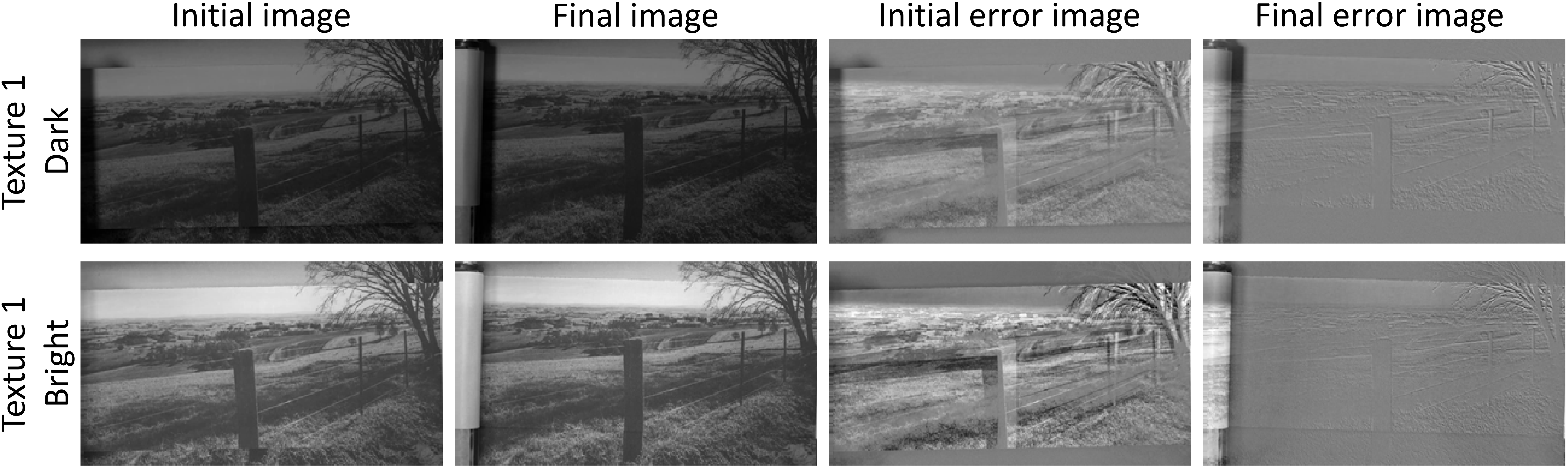}
\caption{Results of visual servoing for Texture~1 under varying lighting conditions (dark and bright).\label{FIG_ROBUSTNESS}}
\end{figure}

To evaluate the generalization capability of the network under varying lighting conditions, experiments using Texture~1 are conducted under both dark and bright lighting conditions. The results are shown in Fig.~\ref{FIG_ROBUSTNESS}. Despite the presence of shadows and partial occlusions caused by the end-effectors, the proposed control scheme maintains stable convergence behavior. These results demonstrate the robustness of the proposed method against environmental disturbances such as lighting variations and occlusions.

\subsubsection{Visual Servoing of Repetitive Texture with Wrinkles}
To further evaluate the robustness of the proposed method, we conducted an additional experiment using a fabric that simultaneously exhibits a repetitive texture pattern and wrinkles (i.e., crumpled-like deformations) under a sagging configuration. In this experiment, the network was trained with RGB inputs to verify that the proposed framework can also operate with color information. Note that the RGB-trained network achieved approximately 14\% lower test loss compared to the grayscale version. As shown in the supplementary video, the proposed method achieved stable convergence while aligning the repetitive texture, even in the presence of deformations and wrinkles. These results further show that the proposed approach generalizes well to realistic garment manipulation scenarios involving non-flat surfaces and repetitive textures.

\subsubsection{Discussion}
\begin{figure}[t]
    \centering
    \subfigure[Texture~1]{
        \includegraphics[width=38mm]{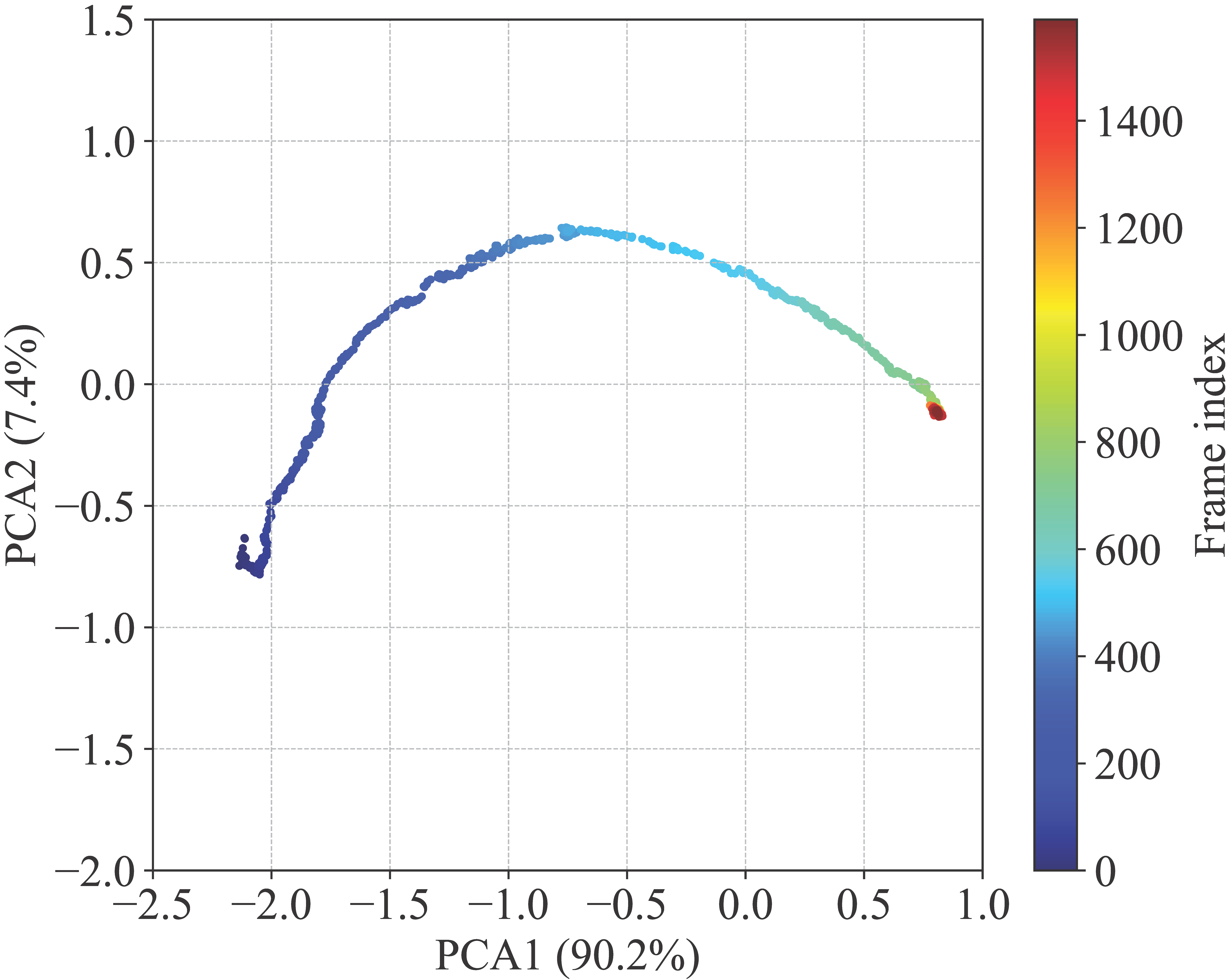}
        \label{PCA1}
    }
    \subfigure[Texture~1 to 5]{
        \includegraphics[width=40mm]{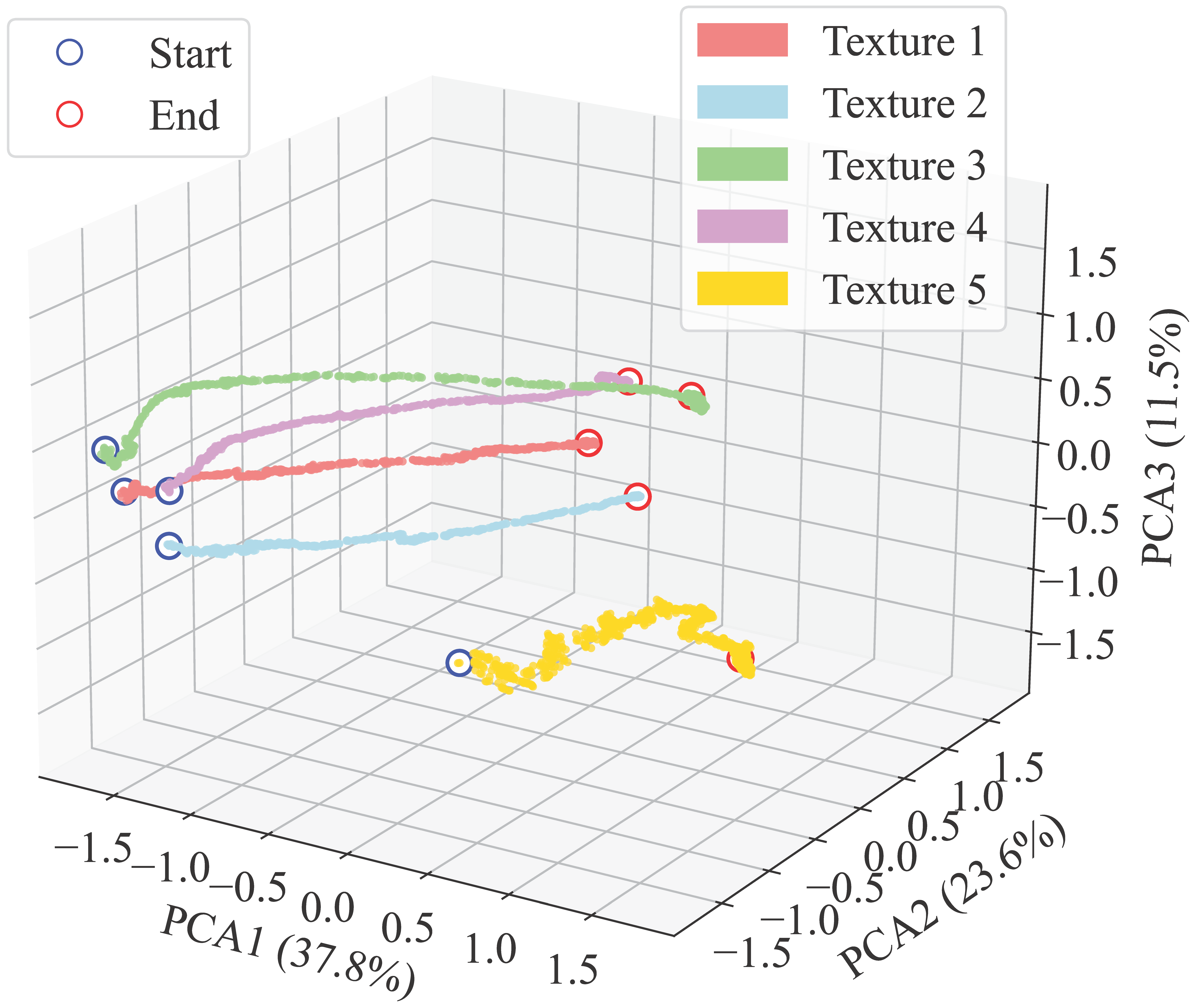}
        \label{PCA2}
    }\\
    \caption{Principal component analysis (PCA) of image features during visual servoing. (a) Features for Texture~1 in PCA space. (b) Features for Textures~1 to 5 in PCA space.\label{PCA}}
\end{figure}

To gain insight into the feature representations of the network, we perform principal component analysis (PCA) on the intermediate features extracted from the global average pooling (GAP) layer. Fig.~\ref{PCA}~\subref{PCA1} shows the PCA results of features extracted during visual servoing with Texture~1, based on the experiments in Fig.~\ref{FIG_TEXTURE1_RESULTS}. The extracted features form a continuous trajectory from the start to the end of visual servoing in the PCA space, suggesting that the network has learned a well-structured feature representation.

Fig.~\ref{PCA}~\subref{PCA2} shows the PCA results of features extracted during visual servoing with all five textures, based on the experiments in Fig.~\ref{FIG_TEXTURE_ALL_RESULTS}. The features corresponding to each texture form a continuous trajectory in the feature space, demonstrating the generalization capability of the proposed network to textures not included in the training dataset. The clear separability between features of different textures further indicates that the network effectively captures texture-specific differences, which is essential for achieving robust visual servoing across previously unseen textures.

A notable observation is the less consistent trajectory of Texture~5, which contains sparse patterns and a low-contrast appearance, making feature extraction more challenging. In contrast, textures with rich visual details and strong contrast, such as Texture~1 to Texture~4, exhibit more consistent trajectories. These results indicate that the network learns a structured and discriminative feature space that enables robust visual servoing across diverse texture types. While low-saliency and low-contrast textures remain more challenging, performance may be further improved by incorporating additional visual cues, such as structured light.

\section{CONCLUSION} 
This paper proposes a method to align and place a fabric piece on another using a dual-arm manipulator and a grayscale camera, enabling accurate texture matching. The system combines Transformer-based visual servoing and dual-arm impedance control to control pose while keeping the fabric piece flat. Trained solely on synthetic images, the network enables zero-shot generalization to real fabrics. Experiments confirm accurate alignment across diverse textures. Code is available at: \url{https://github.com/tfductft/paper-code.git}.


\bibliographystyle{IEEEtran}
\bibliography{./IEEEabrv,./reference}

@ARTICLE{kobayashi2025rollup,
  author={Kobayashi, Akinari and Dong, Wenbo and Seino, Akira and Tokuda, Fuyuki and Kosuge, Kazuhiro},
  journal={IEEE Robot. Autom. Lett.}, 
  title={RollUP: Rolling-Up End-Effector for Fabric Handling}, 
  year={2025},
  volume={10},
  number={11},
  pages={11204--11211}}

@article{F.Chaumette,
  title={Visual servo control. I. Basic approaches},
  author={Chaumette, Fran{\c{c}}ois and Hutchinson, Seth},
  journal={IEEE Robot. Autom. Mag.},
  volume={13},
  number={4},
  pages={82--90},
  year={2006},
}

@article{C.Collewet,
  title={Photometric visual servoing},
  author={Collewet, Christophe and Marchand, Eric},
  journal={IEEE Trans. Robot.},
  volume={27},
  number={4},
  pages={828--834},
  year={2011}
}

@inproceedings{Saxena,
  title={Exploring convolutional networks for end-to-end visual servoing},
  author={Saxena, Aseem and Pandya, Harit and Kumar, Gourav and Gaud, Ayush and Krishna, K Madhava},
  booktitle={Proc. IEEE Int. Conf. Robot. Autom.},
  pages={3817--3823},
  year={2017}
}

@inproceedings{Bateux,
  title={Training deep neural networks for visual servoing},
  author={Bateux, Quentin and Marchand, Eric and Leitner, J{\"u}rgen and Chaumette, Fran{\c{c}}ois and Corke, Peter},
  booktitle={Proc. IEEE Int. Conf. Robot. Autom.},
  pages={3307--3314},
  year={2018}
}

@inproceedings{Yu,
  title={Siamese convolutional neural network for sub-millimeter-accurate camera pose estimation and visual servoing},
  author={Yu, Cunjun and Cai, Zhongang and Pham, Hung and Pham, Quang-Cuong},
  booktitle={Proc. IEEE/RSJ Int. Conf. Intell. Robots Syst.},
  pages={935--941},
  year={2019}
}

@article{Tokuda,
author = {Tokuda, Fuyuki and Arai, Shogo and Kosuge, Kazuhiro},
year = {2019},
month = {01},
pages = {717-725},
title = {Object Positioning by Visual Servoing Based on Deep Learning},
volume = {55},
journal = {Trans. Soc. Instrum. Control Eng.},
}

@inproceedings{Harish,
  title={Dfvs: Deep flow guided scene agnostic image based visual servoing},
  author={Harish, YVS and Pandya, Harit and Gaud, Ayush and Terupally, Shreya and Shankar, Sai and Krishna, K Madhava},
  booktitle={Proc. IEEE Int. Conf. Robot. Autom.},
  pages={9000--9006},
  year={2020}}

@article{Tokuda2,
  title={Convolutional neural network-based visual servoing for eye-to-hand manipulator},
  author={Tokuda, Fuyuki and Arai, Shogo and Kosuge, Kazuhiro},
  journal={IEEE Access},
  volume={9},
  pages={91820--91835},
  year={2021}
}

@inproceedings{Tokuda3,
  title={CNN-based Visual Servoing for Simultaneous Positioning and Flattening of Soft Fabric Parts},
  author={Tokuda, Fuyuki and Seino, Akira and Kobayashi, Akinari and Kosuge, Kazuhiro},
  booktitle={2023 IEEE International Conference on Robotics and Automation (ICRA)},
  pages={748--754},
  year={2023}}

@article{qi2021contour,
  title={Contour moments based manipulation of composite rigid-deformable objects with finite time model estimation and shape/position control},
  author={Qi, Jiaming and Ma, Guangfu and Zhu, Jihong and Zhou, Peng and Lyu, Yueyong and Zhang, Haibo and Navarro-Alarcon, David},
  journal={IEEE/ASME Trans. Mechatronics},
  volume={27},
  number={5},
  pages={2985--2996},
  year={2021}
}

@article{shetab2023lattice,
  title={Lattice-based shape tracking and servoing of elastic objects},
  author={Shetab-Bushehri, Mohammadreza and Aranda, Miguel and Mezouar, Youcef and {\"O}zg{\"u}r, Erol},
  journal={IEEE Trans. Robot.},
  volume={40},
  pages={364--381},
  year={2023}
}

@article{chi2023diffusion,
  title={Diffusion policy: Visuomotor policy learning via action diffusion},
  author={Chi, Cheng and Xu, Zhenjia and Feng, Siyuan and Cousineau, Eric and Du, Yilun and Burchfiel, Benjamin and Tedrake, Russ and Song, Shuran},
  journal={Int. J. Robot. Res.},
    volume = {44},
    number = {10--11},
    pages = {1684-1704},
    year = {2025},
}

@article{chen2024trakdis,
  title={Trakdis: A transformer-based knowledge distillation approach for visual reinforcement learning with application to cloth manipulation},
  author={Chen, Wei and Rojas, Nicolas},
  journal={IEEE Robot. Autom. Lett.},
  volume={9},
  number={3},
  pages={2455--2462},
  year={2024}
}

@article{wang2024efficient,
  title={Efficient Planar Fabric Repositioning: Deformation-Aware RRT* for Non-prehensile Fabric Manipulation},
  author={Wang, Yunhai and Yang, Lei and Zhou, Peng and Qi, Jiaming and Lu, Liang and Zhu, Jihong and Pan, Jia},
  journal={IEEE Robot. Autom. Lett.},
  year={2024},
  volume={9},
  number={12},
  pages={11258--11265},
}

@inproceedings{tan2019efficientnet,
  title={Efficientnet: Rethinking model scaling for convolutional neural networks},
  author={Tan, Mingxing and Le, Quoc},
  booktitle={Proc. Int. Conf. Mach. Learn.},
  volume = {97},
  pages={6105--6114},
  year={2019}
}

@inproceedings{Chen,
  title={Dynamic convolution: Attention over convolution kernels},
  author={Chen, Yinpeng and Dai, Xiyang and Liu, Mengchen and Chen, Dongdong and Yuan, Lu and Liu, Zicheng},
  booktitle={Proc. IEEE/CVF Conf. Comput. Vis. Pattern Recognit.},
  pages={11030--11039},
  year={2020}
}

@inproceedings{woo2023convnext,
  title={Convnext v2: Co-designing and scaling convnets with masked autoencoders},
  author={Woo, Sanghyun and Debnath, Shoubhik and Hu, Ronghang and Chen, Xinlei and Liu, Zhuang and Kweon, In So and Xie, Saining},
  booktitle={Proc. IEEE/CVF Conf. Comput. Vis. Pattern Recognit.},
  pages={16133--16142},
  year={2023}
}

@article{Mehta1,
  title={Mobilevit: light-weight, general-purpose, and mobile-friendly vision transformer},
  author={Mehta, Sachin and Rastegari, Mohammad},
  journal={arXiv preprint arXiv:2110.02178},
  year={2021}
}

@article{Mehta2,
  title={Separable self-attention for mobile vision transformers},
  author={Mehta, Sachin and Rastegari, Mohammad},
  journal={arXiv preprint arXiv:2206.02680},
  year={2022}
}

@inproceedings{Vaswani,
 author = {Vaswani, Ashish and Shazeer, Noam and Parmar, Niki and Uszkoreit, Jakob and Jones, Llion and Gomez, Aidan N and Kaiser, \L ukasz and Polosukhin, Illia},
 booktitle = {Adv. Neural Inf. Process. Syst.},
 title = {Attention is All you Need},
 volume = {30},
 year = {2017}
}

@inproceedings{wang2020eca,
  title={ECA-Net: Efficient channel attention for deep convolutional neural networks},
  author={Wang, Qilong and Wu, Banggu and Zhu, Pengfei and Li, Peihua and Zuo, Wangmeng and Hu, Qinghua},
  booktitle={Proc. IEEE/CVF Conf. Comput. Vis. Pattern Recognit.},
  pages={11534--11542},
  year={2020}
}

@inproceedings{kosuge1997coordinated,
  title={Coordinated motion control of multiple robots manipulating a large object},
  author={Kosuge, Kazuhiro and Hashimoto, Satoshi and Takeo, Koji},
  booktitle={Proc. IEEE/RSJ Int. Conf. Intell. Robots Syst.},
  volume={1},
  pages={208--213},
  year={1997},
}

@inproceedings{Kosuge4,
  title={Coordinated motion control of dual manipulators for handling a rigid object with non-negligible deformation},
  author={Kosuge, Kazuhiro and Kamei, Kentaro and Nammoto, Takashi},
  booktitle={Proc. IEEE Int. Conf. Robot. Autom.},
  pages={5145--5151},
  year={2014}
}

@article{dosovitskiy2020image,
  title={An image is worth 16x16 words: Transformers for image recognition at scale},
  author={Dosovitskiy, Alexey and Beyer, Lucas and Kolesnikov, Alexander and Weissenborn, Dirk and Zhai, Xiaohua and Unterthiner, Thomas and Dehghani, Mostafa and Minderer, Matthias and Heigold, Georg and Gelly, Sylvain and others},
  journal={arXiv preprint arXiv:2010.11929},
  year={2020}
}

@article{loshchilov2017decoupled,
  title={Decoupled weight decay regularization},
  author={Loshchilov, Ilya and Hutter, Frank},
  journal={arXiv preprint arXiv:1711.05101},
  year={2017}
}

@article{Polyak,
  title={Acceleration of stochastic approximation by averaging},
  author={Polyak, Boris T and Juditsky, Anatoli B},
  journal={SIAM J. Control Optim.},
  volume={30},
  number={4},
  pages={838--855},
  year={1992}
}

@inproceedings{hu2018squeeze,
  title={Squeeze-and-excitation networks},
  author={Hu, Jie and Shen, Li and Sun, Gang},
  booktitle={Proc. IEEE/CVF Conf. Comput. Vis. Pattern Recognit.},
  pages={7132--7141},
  year={2018}
}









\end{document}